\documentclass[]{TEAI}
\usepackage{helvet}

\usepackage{amsmath} 
\usepackage{natbib}
\usepackage{graphicx}
\usepackage{subcaption} 
\usepackage{todonotes}
\usepackage[toc,page,header]{appendix}
\usepackage[utf8]{inputenc} 
\usepackage[T1]{fontenc}    
\usepackage{hyperref}       
\usepackage{url}            
\usepackage{booktabs}       
\usepackage{amsfonts}       
\usepackage{nicefrac}       
\usepackage{microtype}      
\usepackage{wrapfig}

\usepackage{amssymb}  
\usepackage{fontawesome}  
\usepackage{url}  

\usepackage{titletoc}

\usepackage{tikz}  
\usepackage{comment}  
\usepackage{tabularx}  
\usepackage{booktabs}  

\usepackage{minitoc}

\usepackage{booktabs}
\usepackage{array}
\usepackage{etoolbox}

\definecolor{lightblue}{RGB}{200, 230, 255}  
\definecolor{headerblue}{RGB}{150, 200, 255} 

\usepackage{pgfplots}
\usepackage[utf8]{inputenc} 
\usepackage[T1]{fontenc}    
\usepackage{hyperref}       
\usepackage{url}            
\usepackage{booktabs}       
\usepackage{amsfonts}       
\usepackage{nicefrac}       
\usepackage{microtype}      
\usepackage{xcolor}         
\usepackage{graphicx}
\usepackage{float}
\usepackage{comment}
\usepackage{multirow} 
\usepackage{amsmath} 
\usepackage{makecell} 
\usepackage{siunitx}  
\usepackage{tikz}
\usepackage{pgf-pie} 
\usepackage{subcaption}
\usepackage{wrapfig}
\usepackage[export]{adjustbox}

\usepackage{amsfonts,amssymb}
\usepackage{ragged2e}      
\usepackage{tabularx}       
\usepackage{array}          
\usepackage{caption}        
\usepackage{enumitem}
\usepackage{pifont}
\usepackage[hang,flushmargin]{footmisc} 

\usepackage{tcolorbox}

\usepackage{tcolorbox}    
\tcbuselibrary{breakable}  
\tcbuselibrary{skins}      

\usepackage{tabularx}
\usepackage{listings}

\DeclareMathAlphabet{\mathcal}{OMS}{cmsy}{m}{n}

\usepackage{tikz}
\usetikzlibrary{calc}

\definecolor{heatup}{RGB}{103,194,172}    
\definecolor{heatdown}{RGB}{235,115,115}  
\definecolor{heatbase}{gray}{0.92}        

\newcommand{\heatdelta}[2]{%
\begingroup
\def\diff{#2}
\ifx\diff\empty
    \begin{tikzpicture}[baseline=-0.6ex]
      \fill[heatbase, rounded corners=0.6] (-2.6em, -1.2ex) rectangle (2.6em, 1.5ex);
      \node[inner sep=0pt, font=\scriptsize, anchor=center] at (0, 0.1ex) {\textbf{#1}};
    \end{tikzpicture}%
\else
    \pgfmathsetmacro{\ispos}{\diff > 0 ? 1 : 0}
    \pgfmathsetmacro{\intensity}{min(abs(\diff)/2.2 * 85 + 15, 100)}
    
    \ifnum\ispos=1
        \begin{tikzpicture}[baseline=-0.6ex]
          \fill[heatup!\intensity!white, rounded corners=0.6] (-2.6em, -1.2ex) rectangle (2.6em, 1.5ex);
          \node[inner sep=0pt, font=\scriptsize, anchor=center] at (0, 0.1ex) {\textbf{#1}\textcolor{black!65}{\tiny~(#2)}};
        \end{tikzpicture}%
    \else
        \begin{tikzpicture}[baseline=-0.6ex]
          \fill[heatdown!\intensity!white, rounded corners=0.6] (-2.6em, -1.2ex) rectangle (2.6em, 1.5ex);
          \node[inner sep=0pt, font=\scriptsize, anchor=center] at (0, 0.1ex) {\textbf{#1}\textcolor{black!65}{\tiny~(#2)}};
        \end{tikzpicture}%
    \fi
\fi
\endgroup
}

\newcommand{\colorbardiverging}{
\begin{tikzpicture}[baseline=(current bounding box.center)]
  \shade[bottom color=heatdown!100, top color=white] (0,0) rectangle (0.3, 1.6);
  \shade[bottom color=white, top color=heatup!100] (0,1.6) rectangle (0.3, 3.2);
  
  \foreach \y/\label in {0/-2.2, 0.8/-1.1, 1.6/0, 2.4/+1.1, 3.2/+2.2}{
    \node[right] at (0.35, \y) {\tiny \label};
  }
  \node[rotate=90, above, font=\scriptsize] at (-0.15, 1.6) {$\Delta$ Accuracy};
\end{tikzpicture}
}

\title{Steering the Verifiability of Multimodal AI Hallucinations}

\author{
    Jianhong Pang\textsuperscript{1,2},
    Ruoxi Cheng\textsuperscript{1,2},
    Ziyi Ye\textsuperscript{1,2,$\dagger$},
    Xingjun Ma\textsuperscript{1,2},\\
    Zuxuan Wu\textsuperscript{1,2},
    Xuanjing Huang\textsuperscript{1,2},
    Yu-Gang Jiang\textsuperscript{1,2}
}

\affiliation[1]{\mbox{Institute of Trustworthy Embodied AI, Fudan University}} 
\affiliation[2]{\mbox{Shanghai Key Laboratory of Multimodal Embodied AI}} 

\abstract{
\begin{abstract}

AI applications driven by multimodal large language models (MLLMs) are prone to hallucinations and pose considerable risks to human users. 
Crucially, such \textbf{hallucinations are not equally problematic}: some hallucination contents could be detected by human users~(i.e., \textbf{obvious hallucinations}), while others are often missed or require more verification effort~(i.e., \textbf{elusive hallucinations}).
This indicates that multimodal AI hallucinations vary significantly in their verifiability. 
Yet, little research has explored how to control this property for AI applications with diverse security and usability demands.
To address this gap, we construct a dataset from 4,470 human responses to AI-generated hallucinations and categorize these hallucinations into obvious and elusive types based on their verifiability by human users.
Further, we propose an activation-space intervention method that learns separate probes for obvious and elusive hallucinations. 
We reveal that obvious and elusive hallucinations elicit different intervention probes, allowing for fine-grained control over the model's verifiability. 
Empirical results demonstrate the efficacy of this approach and show that targeted interventions yield superior performance in regulating corresponding verifiability.
Moreover, simply mixing these interventions enables flexible control over the verifiability required for different scenarios.

\end{abstract}
}

\checkdata[Code]{\url{https://github.com/pang-jh/Steering_the_Verifiability}}
\checkdata[Dataset]{\url{https://huggingface.co/datasets/BeEnough/HHVD}}

\begin{document}
\maketitle
\renewcommand{\thefootnote}{}
\footnotetext{$^\dagger$Corresponding authors.}
\footnotetext{Authors' email addresses: Jianhong Pang(\url{jhpang25@m.fudan.edu.cn}), Ruoxi Cheng(\url{rosycheng12@gmail.com}), Ziyi Ye(\url{zyye@fudan.edu.cn}), Xingjun Ma(\url{xingjunma@fudan.edu.cn}), Zuxuan Wu(\url{zxwu@fudan.edu.cn}), Xuanjing Huang(\url{xjhuang@fudan.edu.cn}), Yu-Gang Jiang(\url{ygj@fudan.edu.cn}).} 
\renewcommand{\thefootnote}{\arabic{footnote}}


\vspace{-1.5em}

\section{Introduction}

Multimodal AI driven by multimodal large language models~(MLLMs) have shown impressive capability in understanding both visual and textual content and generating fluent responses~\cite{liu2023visual}. 
However, MLLMs are known to suffer from hallucinations\cite{guan2024hallusionbench,bai2024hallucination,liu2024survey,wang2023amber}, which means the models generate responses that do not align with the corresponding visual content. 
This problem has attracted growing attention as it poses considerable risks when users cannot readily identify hallucinations in model outputs~\cite{zhou2024relying,cohen2024don,steyvers2025large}.
For example, such hallucinations can lead users to internalize incorrect information, leading to cognitive misguidance and harmful downstream decisions, as well as erosion of trust in deployed AI systems.

\begin{wrapfigure}{r}{0.5\textwidth}
    \centering
    \includegraphics[width=0.5\textwidth]{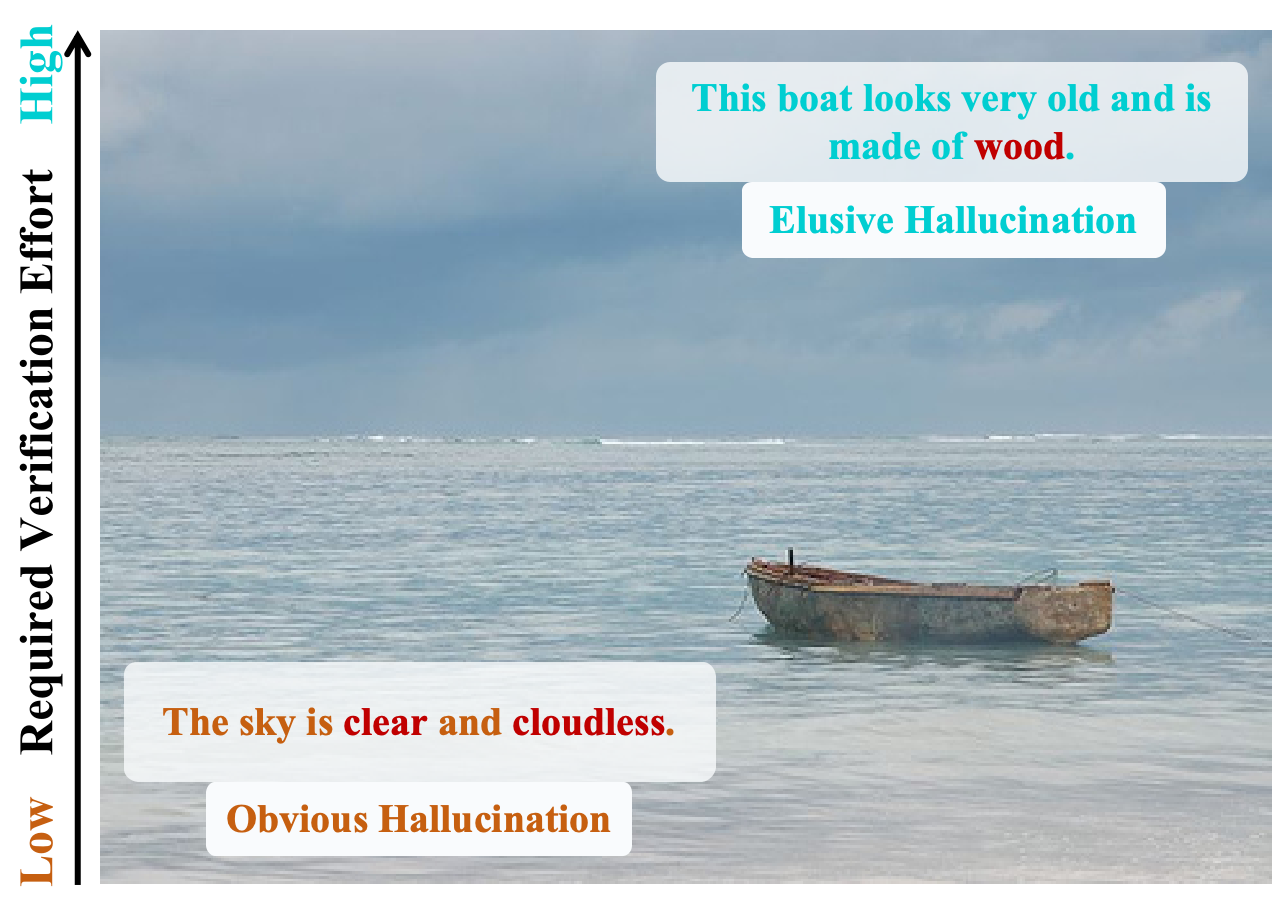}
    \caption{Illustration of verifiability of hallucinations. The same image--text pair can yield hallucinations that are obvious~(easy to verify) or elusive~(difficult to verify).}
    \label{figs:intro}
\end{wrapfigure}

\textbf{Despite existing efforts to mitigate Multimodal AI hallucinations, a key but underexplored aspect is that hallucinations generated by AI are not equally verifiable.}
As illustrated in Figure~\ref{figs:intro}, hallucinations could be obvious~(easy to verify) or elusive~(difficult to verify).
In this example, an obvious hallucination that introduces an incorrect scene-level claim~(``cloudless''), which is easily spotted by users.
While an elusive hallucination that makes a fine-grained attribute claim~(``the boat is made of wood'') is harder to verify without careful inspection, since it concerns a local material property rather than a salient scene-level inconsistency.
In the data construction detailed below, only 20\% of participants successfully identified the error in this elusive case, whereas all participants correctly identified the obvious hallucination.
Notably, the median response time for the elusive case was only 2.3 seconds, suggesting that many users quickly accepted the statement without performing careful verification.
This contrast highlights that hallucinations differ in human's verification effort, and elusive hallucinations can be more likely to mislead the user.

The above example suggests that the impact of hallucinations can differ based on how easily users can recognize and verify them. 
For example, in a human-in-the-loop workflow, obvious hallucinations can have their adverse effects mitigated, while elusive ones may lead to negative consequences. 
Despite substantial advances in hallucination detection and mitigation for MLLMs~\cite{liu2024survey,suresh2025cross,huang2025survey,wang2023amber,gao2025h}, we argue that mitigating hallucinations without accounting for their human verifiability yields an incomplete understanding of the problem.
Currently, there is a lack of (i) a data collection and evaluation framework that stratifies hallucinations by human verifiability, and (ii) methods that can selectively control the verifiability of hallucinations for scenarios with different risks without broadly degrading the general capability of MLLMs.


To address this gap, we study multimodal hallucinations from a user-centric perspective and focus on human verifiability. 
We first construct a human-centered evaluation framework that stratifies hallucinations into obvious and elusive types through a time-constrained annotation protocol.
Specifically, we constructed a dataset of image-text pairs generated by multimodal language models and recruited 40 volunteers to judge their correctness within a 15-second limit for each pair.
Based on 4,470 human responses (five independent judgments per pair), we categorize the dataset based on identification accuracy and response time, resulting in a benchmark of 1,259 samples.
Motivated by recent studies using internal model representations in the activation space to modulate the model's behavior\cite{su2025activation,ji2025calibrating}.
We propose an activation-space intervention method\cite{arditi2024refusal,belrose2023leace} that extracts residual streams related to obvious and elusive hallucination and applies tunable directional ablation with strength $\alpha$ for controlling the models' behavior.
For ease of exposition, we refer to the corresponding type-specific interventions as the \textbf{O}bvious \textbf{H}allucination \textbf{I}ntervention (OHI) and the \textbf{E}lusive \textbf{H}allucination \textbf{I}ntervention (EHI), respectively.
Across three MLLMs, both OHI and EHI reduce hallucination rate and improve accuracy on both \textbf{O}bvious \textbf{H}allucination \textbf{S}ubset (OHS) and the \textbf{E}lusive \textbf{H}allucination \textbf{S}ubset (EHS). 
Importantly, targeted interventions are more effective on their matched subsets, indicating that hallucinations with different human verifiability are associated with distinguishable intervention directions.
For instance, on Qwen2.5-VL-3B, OHI reduces hallucination rate by 32\% on OHS and 25\% on EHS.
At the same time, performance on general benchmarks such as TextVQA only drops by 0.28\% in accuracy. 
Meanwhile, we found that OHI and EHI can be mixed to provide flexible control over hallucination verifiability under scenarios with varying risk and usability.


In summary, our contributions are as follows:

\begin{itemize}
    \item We categorize hallucinations into obvious and elusive types based on the correctness of human responses. Based on this categorization, we construct a dataset with responses from human users.  

    \item We propose an activation-space intervention framework that extracts two hallucination directions (obvious and elusive) from residual-stream representations and applies tunable directional ablation with strength $\alpha$ for fine-grained, disentangled control of verifiability.
    
    \item Extensive experiments across multiple MLLMs show that flexible intervention guided by hallucination patterns distilled from our human-annotated dataset yields markedly different effects on obvious and elusive hallucinations types.
\end{itemize}
\section{Data Construction}

\begin{figure*}[t]
    \centering
    \includegraphics[width=\textwidth]{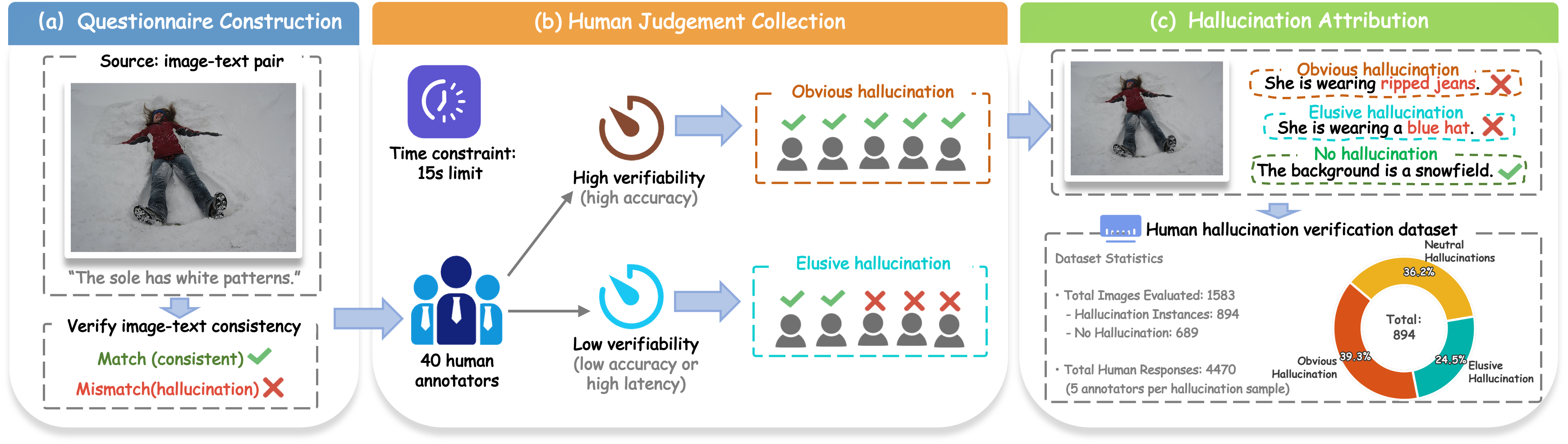}
    \caption{Overview of our data construction pipeline. 
    (a) constructs questionnaires with candidate image-text pairs; (b) collects human judgments; (c) performs hallucination attribution to build a human-labeled dataset.}
    \label{figs:data_construction}
\end{figure*}

This section elaborates on the data construction process. 
Multimodal hallucination refers to cases where the model-generated text contradicts visual facts. 
We distinguish between obvious hallucinations, which involve salient inconsistencies that are easy to verify, and elusive hallucinations, which involve subtle attribute, relation, or knowledge errors that are harder to detect and may pose a higher risk.
Based on this distinction, we construct a human-annotated dataset using questionnaire responses from recruited participants, and use the resulting human-annotated data for training, validation, and testing.
The following subsections describe the construction procedure and dataset statistics. 
Figure~\ref{figs:data_construction} illustrates the overall pipeline.
This study was approved by the Ethics Committee of Fudan University.


\subsection{Questionnaire Construction}

Starting from images collected from the AMBER dataset\cite{wang2023amber}, we first generate raw text descriptions for each image. 
Standard image captioning is often overly concise and conservative, making it difficult to obtain raw text data rich in visual details and potential fine-grained hallucinations. 
To address this limitation, we designed a prompt characterized by exhaustive description and forced inference.

Specifically, we instructed the MLLM to generate detailed image descriptions. 
To ensure both comprehensiveness and depth, our prompt mandates the inclusion of information across three distinct dimensions:
\begin{itemize}
\item \textbf{Panoramic Objects:} Covering not only foreground subjects but also tiny objects in the background;
\item \textbf{Microscopic Attributes:} Including fine-grained details such as materials, specific color codes, OCR text, and exact object counts;
\item \textbf{Spatial Interactions:} Describing relative positions, occlusions, and physical contacts between objects.
\end{itemize}

Crucially, to induce the model to reveal its cognitive boundaries regarding uncertain information, we incorporated a ``bold speculation'' directive into the prompt. 
We prohibited the model from responding with ``unclear'' or ``unknown'', instead requiring it to complete ambiguous details based on scene common sense, thereby eliciting hallucinations. 
This strategy ensures that the generated text maintains an appropriate length and high semantic density, mitigating the information scarcity typical of short texts while preventing the logical incoherence often associated with excessively long generations (see Appendix~\ref{app:prompt construction} for details).

\subsection{Participants}

We recruited forty volunteers (29 males, 11 females, aged 18–60) via social media. All were native Chinese speakers, reducing language-related confounds. 
Participants represented diverse academic backgrounds (e.g., Mathematics, Design, Computer Science), minimizing domain-specific biases associated with any single area.

\subsection{Task Procedure}

Data collection involved multiple online questionnaires, each completed by five independent participants. 
Each item paired an image with a short textual description. 
Candidate image--text pairs and ground truth labels~(i.e., whether hallucinations happened) were constructed with reference to the object-level annotations provided by AMBER~\cite{wang2023amber} when applicable.
And all pairs go through a human verification process with AI experts.
Hallucinated samples contained exactly one word inconsistent with the image. 
To ensure integrity and discourage shortcut strategies, we included fully correct pairs as distractors. 
Participants inspected each pair to identify the hallucinated word, leaving the item blank and clicking ``Next'' if they judged the description entirely consistent.

To approximate rapid, intuitive judgments and to probe potential differences between obvious and elusive hallucinations, we imposed a strict time limit of 15 seconds per item. If a participant did not respond within the time limit, the item was automatically advanced and recorded as a missing response.

\subsection{Dataset Preprocessing}

We operationalize the distinction among obvious, elusive, and neutral hallucinations using annotator-level accuracy and response time aggregated over the five participants. This consensus-based approach reflects the difficulty humans face in perceiving hallucinations. For timing, if a participant does not respond within the 15-second limit, we record the response time as $15$ seconds.
\begin{itemize}
\item \textbf{Obvious Hallucination.} A sample is labeled as obvious if the identification rate is at least $80\%$ (i.e., at least 4/5 participants successfully locate the hallucinated word), indicating that the majority can identify the inconsistency with minimal effort.
\item \textbf{Elusive Hallucination.} A sample is labeled as elusive if the identification rate is at most $40\%$ (i.e., 2/5 participants succeed). In addition, samples with a $60\%$ identification rate (i.e., 3/5 participants succeed) are also labeled as elusive when the median response time across participants exceeds $12$ seconds, suggesting higher verification cost even among successful identifications.
\item \textbf{Neutral Hallucination.} A sample is labeled as neutral if it does not satisfy the criteria for either obvious or elusive hallucination. This category captures hallucinations with intermediate verifiability.
\end{itemize}

\subsection{Data Statistics}

After filtering and categorization, we constructed a final dataset of 1,259 high-quality samples, including 689 non-hallucinated instances, 351 obvious hallucinations, and 219 elusive hallucinations. Each sample is an image--text pair in a discriminative evaluation format: given an image and a description, the model is asked to judge whether the description is consistent with the image by answering only ``Yes'', ``No'', or ``Uncertain''.

To facilitate human annotation, this dataset is composed of short, fine-grained visual statements.
In terms of content, obvious hallucinations involve salient scene-level inconsistencies, such as fabricated objects or clearly mismatched visible elements, whereas elusive hallucinations concern subtle attributes, materials, or localized relations that are more harder to verify at a glance. Non-hallucinated samples consist of image-grounded descriptions without mismatches. 
For three subsets (obvious-hallucination, elusive-hallucination, non-hallucinated), we partition the dataset into training, validation, and test sets using a 55\%:20\%:25\% split. 
This design makes the dataset suitable for evaluating the verifiability of concrete visual claims.

\section{Method}
\subsection{Task Definition}

Let $\mathcal{M}$ be a transformer-based MLLM. Given an image $I$ and a textual instruction $T$, the model generates a response $Y = \{y_1, \ldots, y_n\}$ via autoregressive decoding. Following standard transformer notation, we denote the residual-stream activation
at layer $l$ and token position $i$ as $\mathbf{x}^{(l)}_i \in \mathbb{R}^{d_{\text{model}}}$.

We define post-instruction tokens as the template tokens that appear after the instruction content in the model's chat template. Our analysis focuses on activations in this region, as it captures the stage where the model transitions from processing the prompt to forming its response. The chat templates for all models and the corresponding post-instruction tokens are provided in Appendix~\ref{app:Chat Templates}.

We study multimodal hallucination, where the generated text contradicts image content. Moreover, we consider two risk levels: obvious hallucinations and elusive hallucinations. Our goal is to identify directions in activation space that capture hallucination tendencies, and to intervene on $\mathbf{x}^{(l)}_i$ at inference time to reduce hallucinated generations while preserving the model's capability.

Concretely, we aim to extract two direction vectors, $\mathbf{r}_{\text{oh}}$ and $\mathbf{r}_{\text{eh}}$,
corresponding to obvious and elusive hallucination behaviors, respectively, and apply a controllable intervention that suppresses components aligned with these directions.

\subsection{Difference Vector Selection}

\textbf{Difference-in-means.}
To isolate hallucination-related features, we adopt the difference-in-means technique\cite{belrose2023leace}, which characterizes differences in mean activations between contrasting datasets in residual stream. 
This technique is highly data-efficient and extracts robust feature directions from only a few hundred high-quality contrastive samples.
Let $\mathcal{D}^{(\text{train})}_{\text{nh}}$ denote non-hallucinated samples, and $\mathcal{D}^{(\text{train})}_{\text{type}}$ denote hallucinated samples of a given $\text{type} \in \{\text{oh}, \text{eh}\}$.
For each layer $l$ and post-instruction token position $i$, we compute:

\begin{equation}
\begin{aligned}
\boldsymbol{\mu}^{(l)}_i(\text{type})
&= \frac{1}{|\mathcal{D}^{(\text{train})}_{\text{type}}|}
\sum_{t \in \mathcal{D}^{(\text{train})}_{\text{type}}} \mathbf{x}^{(l)}_i(t), \quad
\boldsymbol{\nu}^{(l)}_i(\text{nh})
&= \frac{1}{|\mathcal{D}^{(\text{train})}_{\text{nh}}|}
\sum_{t \in \mathcal{D}^{(\text{train})}_{\text{nh}}} \mathbf{x}^{(l)}_i(t).
\end{aligned}
\end{equation}
We then define the candidate hallucination direction as:
\begin{equation}
\mathbf{r}^{(l)}_i(\text{type}) = \boldsymbol{\mu}^{(l)}_i(\text{type}) - \boldsymbol{\nu}^{(l)}_i(\text{nh}).
\end{equation}
Intuitively, (i) the direction of $\mathbf{r}^{(l)}_i$ indicates how hallucinated and non-hallucinated activations separate, while (ii) its magnitude reflects their average distance.

\textbf{Selecting a single vector (per type).}
Computing $\mathbf{r}^{(l)}_i(\text{type})$ for all post-instruction token positions $i \in \mathcal{I}$ and layers $l \in L$ yields a candidate set of size $\mathcal{I} \times L$.
We then select a single most effective vector by evaluating each candidate on a hallucination validation set $\mathcal{D}^{(\text{val})}_{\text{type}}$ and a non-hallucinated validation set $\mathcal{D}^{(\text{val})}_{\text{nh}}$. The selection criterion captures two desiderata: (i) hallucination suppression—when the candidate direction is ablated, it should maximally reduce hallucinated behavior on $\mathcal{D}^{(\text{val})}_{\text{type}}$; and (ii) behavior preservation—it should induce minimal changes to the model’s other behaviors, as measured on $\mathcal{D}^{(\text{val})}_{\text{nh}}$.
A more detailed description of our selection algorithm is provided in Appendix~\ref{app:Direction Selection}.
We denote the selected direction as $\mathbf{r}$ and its unit-normalized version as $\hat{\mathbf{r}}$.

\subsection{Model Intervention}
\label{model_intervention}
To investigate the functional role of a unit direction $\hat{\mathbf{r}}_{\text{type}} \in \mathbb{R}^{d_{\text{model}}}$ in the model's computation, we adopt directional ablation, which removes the component of the residual-stream representation aligned with this direction. For any residual activation $\mathbf{x} \in \mathbb{R}^{d_{\text{model}}}$, directional ablation subtracts the orthogonal projection of $\mathbf{x}$ onto $\hat{\mathbf{r}}_{\text{type}}$, thereby eliminating the component aligned with $\hat{\mathbf{r}}_{\text{type}}$ in the residual stream. We further introduce a scalar hyperparameter $\alpha$ to obtain a tunable variant that controls the intervention strength and enables analysis of model behavior under different intensities. We apply:
\begin{equation}
\mathbf{x}' \leftarrow \mathbf{x} - \alpha \,\hat{\mathbf{r}}_{\text{type}} \hat{\mathbf{r}}_{\text{type}}^{\top} \mathbf{x}.
\label{eq:model_intervention}
\end{equation}
where $\alpha \ge 0$ is the ablation strength.

At inference time, we apply Eq.~\eqref{eq:model_intervention} globally to the residual stream across all layers $l$ and the relevant token positions $i$. Concretely, within each layer, we intervene on both the pre-attention residual activation $\mathbf{x}^{(l)}_i$ and the post-attention residual activation $\tilde{\mathbf{x}}^{(l)}_i$, ensuring that the target direction is consistently suppressed throughout the forward pass. 
Building on this global intervention scheme, we further adopt a disentangled control strategy: setting $\hat{\mathbf{r}}_{\text{type}}=\hat{\mathbf{r}}_{\text{oh}}$ with a tuned $\alpha$ primarily targets object and semantic fabrications associated with obvious hallucinations, whereas setting $\hat{\mathbf{r}}_{\text{type}}=\hat{\mathbf{r}}_{\text{eh}}$ with a tuned $\alpha$ primarily targets subtle attribute and relationship deviations associated with elusive hallucinations. By evaluating these two interventions separately, we can empirically assess whether obvious and elusive hallucinations correspond to different internal mechanisms, and select appropriate intervention strengths for different risk levels.

For the two hallucination types, Eq.~\eqref{eq:model_intervention} describes the independent intervention applied to each direction $\hat{\mathbf{r}}_{\text{type}}$. Beyond this type-wise intervention, we formulate verifiability control as a continuous steering between the two type-specific directions in activation space. Specifically, given the unit directions $\hat{\mathbf{r}}_{\text{oh}}$ and $\hat{\mathbf{r}}_{\text{eh}}$, we first construct a mixed direction and then apply directional ablation along it. To isolate the effect of directional composition, we define the mixed direction using a single steering variable $\lambda$:
\begin{equation}
\hat{\mathbf{r}}_{\text{mix}}(\lambda)
= \operatorname{norm}\!\big((1-\lambda)\hat{\mathbf{r}}_{\text{oh}}+\lambda \hat{\mathbf{r}}_{\text{eh}}\big),
\qquad \lambda \in [0,1].
\label{eq:mix_direction}
\end{equation}
We then apply directional ablation along the mixed direction:
\begin{equation}
\mathbf{x}' \leftarrow \mathbf{x} - \alpha \, \hat{\mathbf{r}}_{\text{mix}}(\lambda) \hat{\mathbf{r}}_{\text{mix}}(\lambda)^{\top} \mathbf{x}.
\label{eq:mix_intervention}
\end{equation}
Here, $\lambda$ controls the interpolation between the obvious and elusive directions: when $\lambda=0$, the mixed direction reduces to $\hat{\mathbf{r}}_{\text{oh}}$, and when $\lambda=1$, it reduces to $\hat{\mathbf{r}}_{\text{eh}}$. Intermediate values continuously interpolate between the two directions, while $\alpha$ in Eq.~\eqref{eq:mix_intervention} controls the overall ablation strength along the mixed direction. 
This formulation enables direct steering of hallucination verifiability through a single mixed direction rather than treating the two probes as fully separate interventions.

\begin{figure*}[t]
    \centering
    \includegraphics[width=\textwidth]{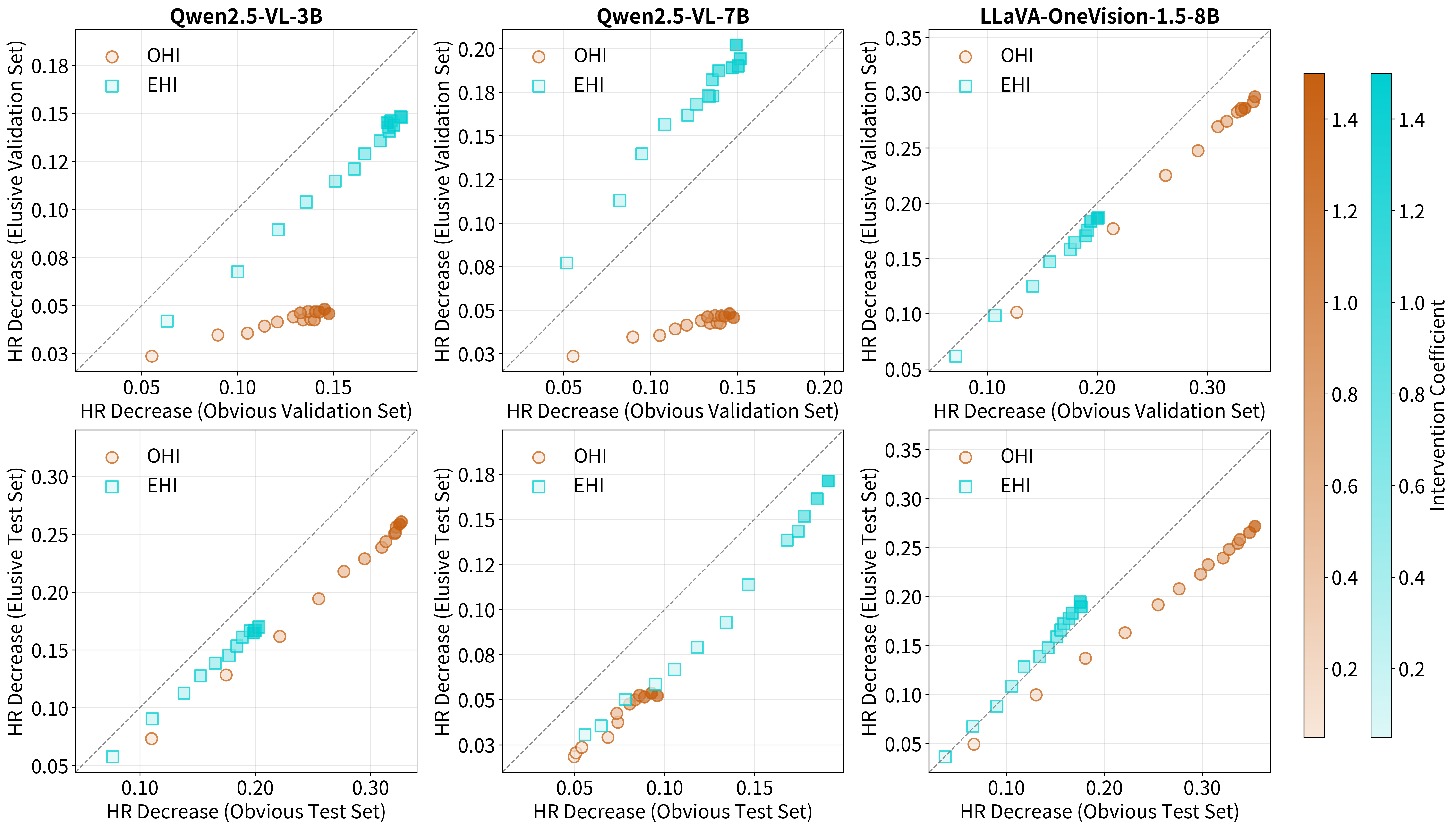}
    \caption{Joint visualization of hallucination-rate reduction under OHI and EHI across validation and test sets. Each point corresponds to a single-direction coefficient sweep, revealing a consistent spatial separation between the two interventions across three MLLMs.}
    \label{figs:obvious_vs_elusive}
\end{figure*}
\section{Experimental Setup}
We evaluate on three open-source MLLMs: \textbf{Qwen2.5-VL-3B}\cite{qwen2.5-VL}, \textbf{Qwen2.5-VL-7B}\cite{qwen2.5-VL}, and \textbf{LLaVA-OneVision-1.5-8B}\cite{an2025llavaonevision15fullyopenframework,xie2025region}. 

\subsection{Performance Evaluation}

\textbf{Evaluation data format.}
To quantify the model's performance in hallucination control, we employed a logits-based discriminative evaluation protocol, where the model is given an image-text pair and required to judge their consistency by answering only “Yes,” “No,” or “Uncertain” (see Appendix~\ref{app:prompt evaluation} for the prompt).

\textbf{Logit-based metrics.}
We normalized the predicted probabilities of these three target tokens, denoted as $P(\text{Yes})$, $P(\text{No})$, $P(\text{Unc})$. We report three core metrics:
\begin{itemize}
    \item \textbf{Hallucination Rate (HR)}: the probability of selecting the incorrect answer. Lower is better.
    
    \item \textbf{Accuracy (ACC)}: the probability of selecting the correct answer. Higher is better.
    
    \item \textbf{Uncertain Tendency (UT)}: the probability of answering ``Uncertain'', measuring conservativeness.
\end{itemize}

\textbf{Comparison of baseline vs. intervention.}
For each model, we first compute the baseline logits distribution without intervention, then recompute logits under intervention. 
Evaluation is conducted on test splits derived from our constructed dataset. 
We report changes (intervention - baseline) and aggregate results by averaging over the test set.

\subsection{General Ability Evaluation}
To assess whether our interventions preserve general-purpose capabilities, we perform two common multimodal evaluations: \textbf{MMBench\_CN}\cite{liu2024mmbench}, and \textbf{TextVQA}\cite{singh2019towards}. All evaluations follow the standard protocol of the multimodal evaluation toolkit VLMEvalKit\cite{duan2024vlmevalkit}.

To further analyze the trade-off between intervention strength and general capability, we conduct a sweep over the intervention coefficient $\alpha$ on \textbf{TextVQA}\cite{singh2019towards}. Since this benchmark relies heavily on fine-grained visual grounding and text recognition, it is typically more sensitive to changes in intermediate representations.

\subsection{Implementation Details}
Experiments are conducted on eight NVIDIA RTX 4090 GPUs. The steering coefficient $\alpha$ used in directional ablation was adjusted dynamically during the inference phase to analyze its impact on model behavior.
\section{Experimental Results}
\subsection{Obvious vs. Elusive Hallucination Intervention}
We investigate the effects of OHI and EHI on MLLM-generated hallucinations with different verifiability, i.e., OHS and EHS.
Figure~\ref{figs:obvious_vs_elusive} shows the intervention results, where each point corresponds to a single-direction intervention with the other direction fixed to zero; darker colors denote larger intervention coefficients.
Across all three MLLMs, most points fall in the upper-right region, showing that both interventions consistently reduce hallucination rate on both subsets. 
In general, directional ablation suppresses inconsistency-driven errors and increases the probability of correct judgments. 

\definecolor{ohicolor}{HTML}{C65F10}
\definecolor{ehicolor}{HTML}{00CED1}
\begin{wraptable}{r}{0.5\textwidth}
\centering
\scriptsize
\setlength{\tabcolsep}{2pt}
\renewcommand{\arraystretch}{1.0}
\caption{Effects of OHI and EHI on ACC(\%) and UT(\%) across different test subsets under the selected intervention settings. Values denote absolute changes relative to the baseline.}
\label{tab:intervention_results_acc_ut}
\begin{tabular}{ll cccc}
\toprule
\multicolumn{1}{c}{\multirow{2.5}{*}{\textbf{Model}}} &
\multicolumn{1}{c}{\multirow{2.5}{*}{\textbf{Type}}} &
\multicolumn{2}{c}{\textbf{OHS}} &
\multicolumn{2}{c}{\textbf{EHS}} \\
\cmidrule(lr){3-4} \cmidrule(lr){5-6}
& & \textbf{ACC(\%)} & \textbf{UT(\%)}
& \textbf{ACC(\%)} & \textbf{UT(\%)} \\
\midrule

\multirow{2}{*}{Qwen2.5-VL-3B}
& \textcolor{ohicolor}{OHI} & \textcolor{ohicolor}{+5.79}  & \textcolor{ohicolor}{+26.28} & \textcolor{ohicolor}{+4.73}  & \textcolor{ohicolor}{+20.3} \\
& \textcolor{ehicolor}{EHI} & \textcolor{ehicolor}{+2.52}  & \textcolor{ehicolor}{+17.10} & \textcolor{ehicolor}{+3.15}  & \textcolor{ehicolor}{+13.50} \\
\midrule

\multirow{2}{*}{Qwen2.5-VL-7B}
& \textcolor{ohicolor}{OHI} & \textcolor{ohicolor}{+9.14}  & \textcolor{ohicolor}{-1.08}  & \textcolor{ohicolor}{+5.22}  & \textcolor{ohicolor}{-0.47} \\
& \textcolor{ehicolor}{EHI} & \textcolor{ehicolor}{+16.99} & \textcolor{ehicolor}{+0.43}  & \textcolor{ehicolor}{+14.22} & \textcolor{ehicolor}{+0.85} \\
\midrule

\multirow{2}{*}{LLaVA-OneVision-1.5-8B}
& \textcolor{ohicolor}{OHI} & \textcolor{ohicolor}{+32.75} & \textcolor{ohicolor}{+2.51}  & \textcolor{ohicolor}{+24.82} & \textcolor{ohicolor}{+1.76} \\
& \textcolor{ehicolor}{EHI} & \textcolor{ehicolor}{+18.51} & \textcolor{ehicolor}{-0.88}  & \textcolor{ehicolor}{+19.75} & \textcolor{ehicolor}{+0.32} \\

\bottomrule
\end{tabular}%
\end{wraptable}

Based on this joint visualization, we further compare the relative effects of OHI and EHI on OHS and EHS separately, and observe a clear spatial separation between the two interventions, with each showing stronger effects on its targeted hallucination subset.
In particular, when OHI and EHI achieve comparable reductions on OHS, EHI yields a substantially larger reduction on EHS. 
Conversely, the obvious probe exhibits a stronger tendency to correct salient, easily detectable errors. 
This robust pattern across different evaluation protocols is consistent with our hypothesis that obvious and elusive hallucinations are encoded in distinct directional subspaces, enabling probe-specific interventions to modulate the two risk types.

We also examine how the interventions affect ACC and UT, as reported in Table~\ref{tab:intervention_results_acc_ut}. 
To make a fair comparison, we report results at the selected intervention setting for each probe. 
Specifically, for OHI and EHI, we select a single intervention coefficient for each by choosing an appropriate intervention strength. 
The chosen strength should achieve a substantial reduction in HR on the validation set while avoiding overly aggressive changes in overall behavior; detailed selection criteria are provided in Appendix~\ref{app:intervention coefficients}. 
Under these selected settings, ACC improves across models and hallucination subsets, indicating that the reduction in hallucination rate is accompanied by more correct judgments rather than merely suppressing model responses. 
At the same time, although UT does increase in some cases, the magnitude of this increase remains limited overall, and the simultaneous gain in ACC indicates that the intervention is still well balanced in practice. 
\begin{wraptable}{r}{0.5\textwidth}
\centering
\scriptsize
\setlength{\tabcolsep}{2pt}
\renewcommand{\arraystretch}{1.0}
\caption{Effects of OHI and EHI on HR(\%) across different test subsets under the selected intervention settings. We also report $\Delta$, defined as the difference between the two subsets ($\Delta = \mathrm{EHI} - \mathrm{OHI}$).}
\label{tab:intervention_results_hr}
\begin{tabular}{l
>{\color{ohicolor}}c
>{\color{ehicolor}}c
c
>{\color{ohicolor}}c
>{\color{ehicolor}}c
c}
\toprule
\multicolumn{1}{c}{\multirow{2}{*}{\textbf{Model}}} &
\multicolumn{3}{c}{\textbf{OHS}} &
\multicolumn{3}{c}{\textbf{EHS}} \\
\cmidrule(lr){2-4} \cmidrule(lr){5-7}
& \textbf{\textcolor{ohicolor}{OHI}} & \textbf{\textcolor{ehicolor}{EHI}} & \textbf{$\Delta$}
& \textbf{\textcolor{ohicolor}{OHI}} & \textbf{\textcolor{ehicolor}{EHI}} & \textbf{$\Delta$} \\
\midrule
Qwen2.5-VL-3B          & -32.07 & -19.56 & +12.51 & -25.04 & -16.65 & +8.39 \\
Qwen2.5-VL-7B          & -8.06  & -17.42 & -9.36  & -4.76  & -15.08 & -10.32 \\
LLaVA-OneVision-1.5-8B & -35.26 & -17.53 & +17.73 & -26.58 & -19.43 & +7.15 \\
\bottomrule
\end{tabular}
\end{wraptable}
Table~\ref{tab:intervention_results_hr} further reports the corresponding HR changes under the same selected settings.
Generally, in terms of performance differences between OHI and EHI, the EHI/OHI method shows a clear advantage on its targeted dataset.
Notably, OHI yields a larger HR reduction for Qwen2.5‑VL‑3B and LLaVA‑OneVision‑1.5‑8B, while EHI achieves greater HR reduction on Qwen2.5‑VL‑7B.
We attribute this to the models' differing baseline visual capabilities: as verified on rigorous benchmarks such as MMMU-Pro Vision\cite{yue2025mmmu}, Qwen2.5-VL-7B exhibits stronger baseline performance.
OHI enhances the model’s basic reasoning ability, thus bringing larger improvements to weaker models.
In contrast, EHI targets more challenging corner cases with stricter constraints, making it more effective on stronger models.

The complete results, including those on the Non-Hallucination test set, are provided in the Appendix~\ref{app:intervention table}.



\subsection{Influence on General Ability}

Beyond mitigating hallucination risk, we examine the impact of our interventions on general ability.

\begin{table*}[t]
\centering
\caption{Influence on general ability under the selected intervention coefficients.
We report accuracy (\%) on general understanding benchmarks. \textbf{Color intensity reflects the change compare to base model.} (Green for improvements, Red for declines).}
\label{tab:general_ability}

\setlength{\tabcolsep}{4pt}
\renewcommand{\arraystretch}{1.3} 
\begin{minipage}{0.88\textwidth}
\centering
\resizebox{\textwidth}{!}{%
\begin{tabular}{l ccc | ccc}
\toprule
\multicolumn{1}{c}{\multirow{2.5}{*}{\textbf{Model}}} &
\multicolumn{3}{c|}{\textbf{MMBench\_CN}} &
\multicolumn{3}{c}{\textbf{TextVQA\_VAL}} \\
\cmidrule(lr){2-4} \cmidrule(lr){5-7}
& \textbf{Base} & \textbf{OHI} & \textbf{EHI} 
& \textbf{Base} & \textbf{OHI} & \textbf{EHI} \\
\midrule
Qwen2.5-VL-3B
& \heatdelta{84.62}{} & \heatdelta{82.85}{-1.77} & \heatdelta{82.42}{-2.20}
& \heatdelta{84.09}{} & \heatdelta{83.81}{-0.28} & \heatdelta{83.45}{-0.64} \\

Qwen2.5-VL-7B
& \heatdelta{86.84}{} & \heatdelta{86.65}{-0.19} & \heatdelta{85.33}{-1.51}
& \heatdelta{89.74}{} & \heatdelta{88.18}{-1.56} & \heatdelta{89.13}{-0.61} \\

LLaVA-OneVision-1.5-8B
& \heatdelta{88.32}{} & \heatdelta{88.67}{+0.35} & \heatdelta{88.27}{-0.05}
& \heatdelta{85.99}{} & \heatdelta{86.11}{+0.12} & \heatdelta{85.93}{-0.06} \\
\bottomrule
\end{tabular}
}
\end{minipage}
\hfill
\begin{minipage}{0.08\textwidth}
\centering
\resizebox{!}{1.5cm}{\colorbardiverging}
\end{minipage}
\end{table*}

Table~\ref{tab:general_ability} reports results on two standard multimodal benchmarks (MMBench\_CN and TextVQA) for three models.
For each benchmark, we compare the baseline performance with performance under the OHI and EHI, and report the absolute change in parentheses relative to the baseline.
Overall, under the selected intervention settings, the performance changes on these benchmarks are modest in most cases, suggesting that hallucination-oriented directional ablation can often be applied without substantially degrading general capability. Additional coefficient-sensitivity analysis on TextVQA is provided in Appendix~\ref{app:alpha_sweep}.
\begin{wrapfigure}{r}{0.45\textwidth}
    \centering
    \includegraphics[width=0.45\textwidth]{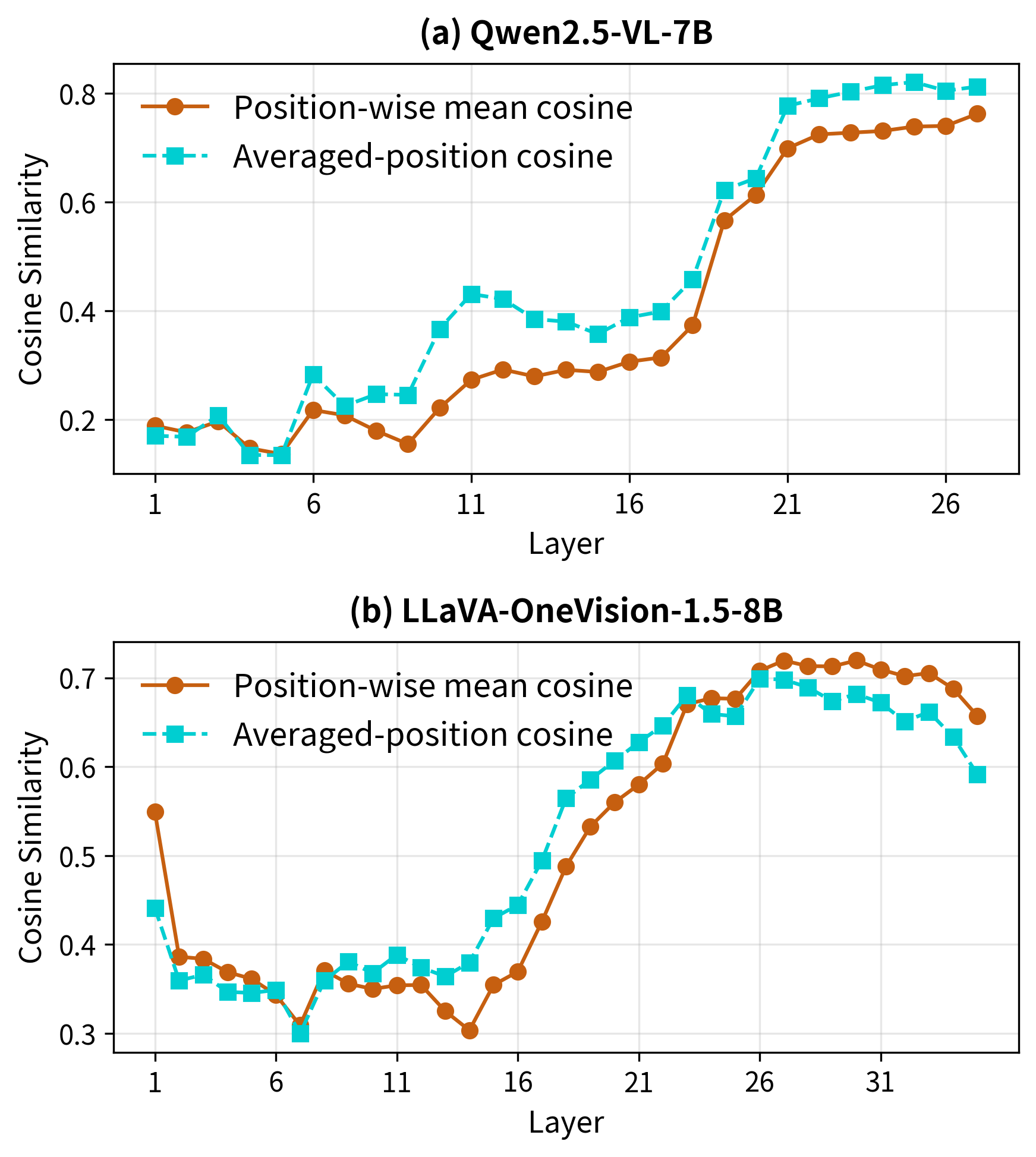}
    \caption{Layer-wise cosine similarity between obvious and elusive hallucination directions.}
    \label{figs:cos_sim}
\end{wrapfigure}
\subsection{Transformer Layer Analysis}

\begin{wrapfigure}{r}{0.5\textwidth}
    \centering
    \includegraphics[width=0.5\textwidth]{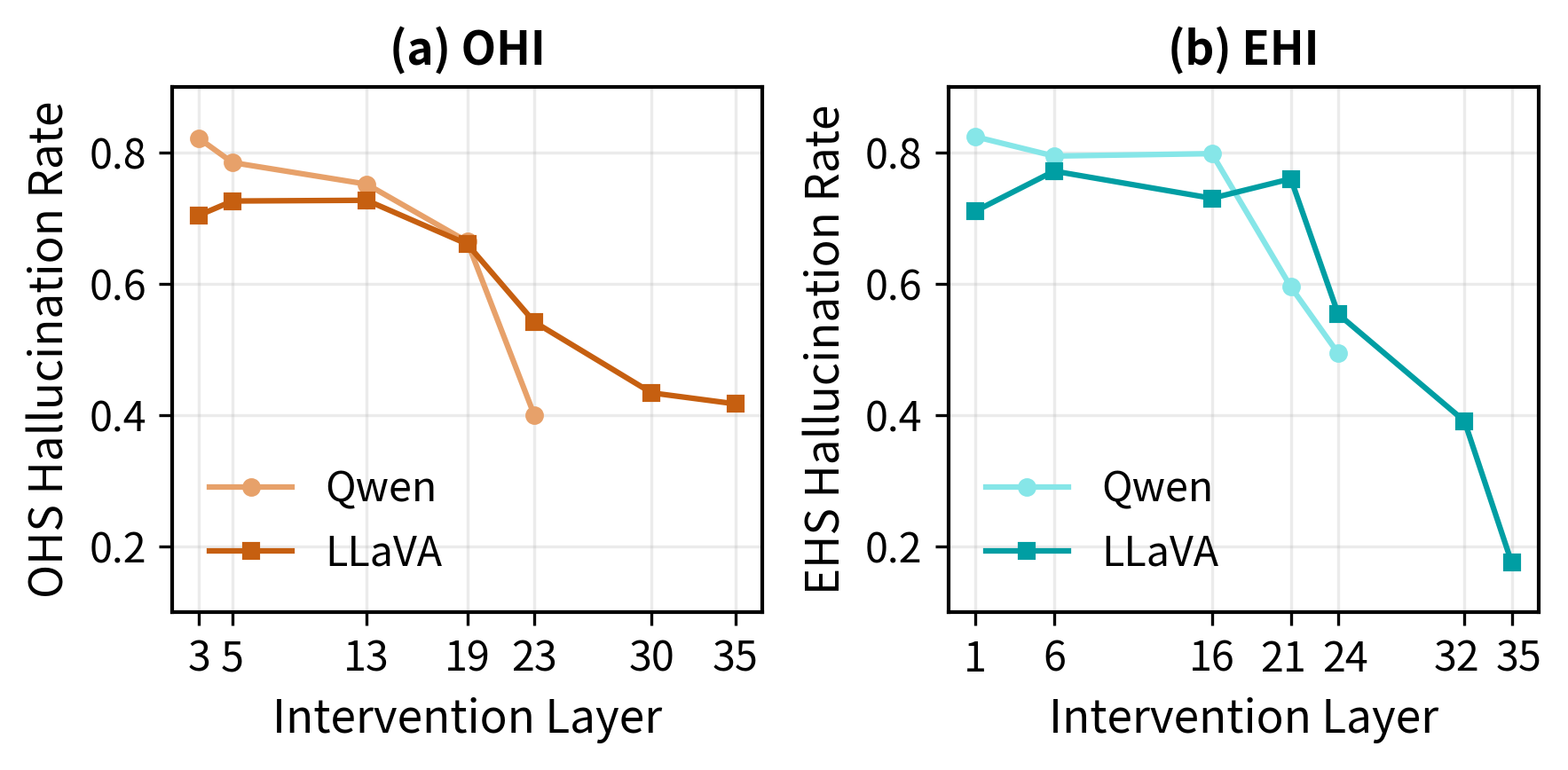}
    \caption{Layer comparison of OHI and EHI in Qwen2.5-VL-7B and LLaVA-OneVision-1.5-8B.
    The figure shows the average hallucination rate over post-instruction token positions.}
    \label{figs:layer_analysis}
\end{wrapfigure}

To understand where hallucination-related representations emerge and how intervention effectiveness varies with depth, we conduct a layer-wise analysis on the models shown in Figure~\ref{figs:layer_analysis}. We apply intervention using features extracted from different layers. 
As shown in Figure~\ref{figs:layer_analysis}, hallucination-related behaviors are not uniformly distributed across layers.
In both models, the most pronounced reductions generally appear in middle-to-late layers rather than in the earlier layers, suggesting that hallucination-relevant features are more strongly represented in deeper parts of the network. We also observe that OHI and EHI exhibit distinct layer-wise patterns.

To further probe their relationship, Figure~\ref{figs:cos_sim} reports the layer-wise cosine similarity between obvious and elusive hallucination directions. The similarity is relatively low in earlier layers but increases steadily in deeper layers, indicating that the two directions become more aligned as depth increases. This suggests that obvious and elusive hallucinations may rely on partially similar representations in deeper layers, while preserving type-specific differences that enable selective intervention in the lower layers.

\subsection{Mixed Interventions and Case Study}

\begin{wrapfigure}{r}{0.5\textwidth}
    \centering
    \includegraphics[width=0.5\textwidth]{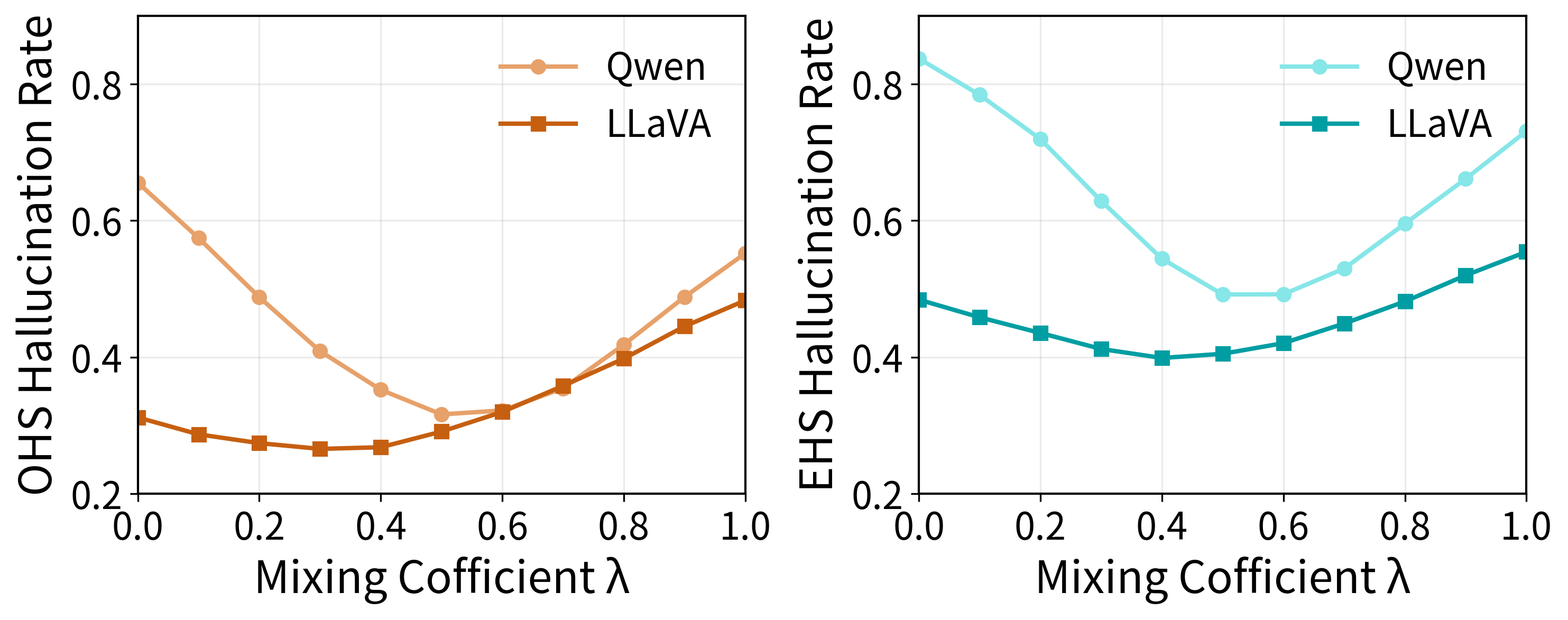}
    \caption{Verifiability steering via the mixed direction on Qwen2.5-VL-7B and LLaVA-OneVision-1.5-8B. We fix the ablation strength $\alpha=1$ and vary the mixing coefficient $\lambda$.}
    \label{figs:mix}
\end{wrapfigure}

We use qualitative examples to illustrate how different intervention directions affect verification behavior at the instance level, while also connecting to the continuous steering effect of the mixed direction. 
To verify that hallucination verifiability can be steered continuously, we evaluate the mixed-direction intervention introduced in Section~\ref{model_intervention}. 
We set the ablation strength $\alpha=1$ and vary only the steering coefficient $\lambda \in [0,1]$. 
As shown in Figure~\ref{figs:mix}, both the obvious hallucination rate and the elusive hallucination rate vary with $\lambda$. Rather than reaching their best values at either endpoint, all curves are lower in the middle range, suggesting that neither a purely obvious-oriented nor a purely elusive-oriented intervention is optimal. Instead, an intermediate mixed direction yields a better trade-off.

\begin{figure*}[t]
    \centering
    \includegraphics[width=\textwidth]{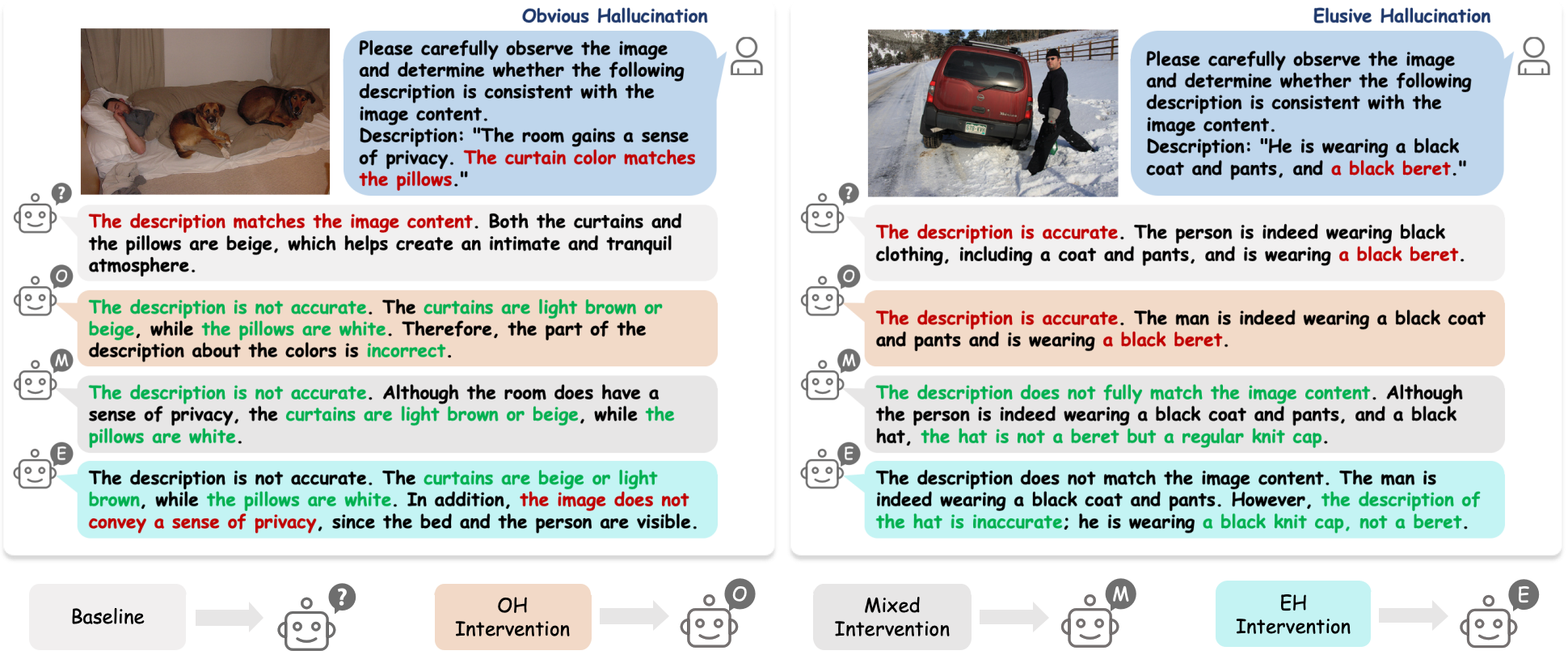}
    \caption{Case studies of intervention effects on LLaVA-OneVision-1.5-8B.
    Hallucinated content is highlighted in red, and corrected or image-grounded content is highlighted in green.
    The mixed-direction intervention is performed with $\lambda = 0.5$.}
    \label{figs:case_study}
\end{figure*}

Figure~\ref{figs:case_study} provides two representative examples under the baseline model, the OHI, the mixed intervention, and the EHI. In the mixed setting, we use an equal combination of the two directions ($\lambda=0.5$).

In the left example, the error is obvious: the description claims that the curtain color matches the pillows, although the curtains are beige and the pillows are white. The baseline model incorrectly accepts the description. The OHI correctly rejects it by focusing on the salient mismatch. The EHI also rejects it, but becomes more overly meticulous by further questioning whether the image conveys a ``sense of privacy.'' The mixed intervention lies between the two, correcting the main inconsistency while avoiding the over-analysis.

In the right example, the hallucination is more elusive: the description states that the person is wearing a black beret, while the hat is more accurately a black knit cap. Both the baseline model and the OHI still accept the description, suggesting that OHI has limited effect on this subtle fine-grained error. In contrast, the EHI successfully identifies the mismatch. The mixed intervention again shows an intermediate behavior, being more sensitive than OHI while less specialized than pure EHI.

Overall, these examples align with the quantitative trend of mixed-direction steering. OHI is more effective for salient and easily verifiable hallucinations, whereas EHI is more sensitive to subtle fine-grained errors but may become overly strict on obvious cases. The mixed intervention provides a compromise between the two, further supporting that hallucination verifiability can be steered between obvious and elusive regimes.

\section{Related Work}

\subsection{Hallucination Detection and Intervention}
Prior work on hallucinations in language models can be broadly categorized into detection and mitigation. 
On the detection side, a common line of research uses model uncertainty signals as indicators of potential hallucinations\cite{farquhar2024detecting,zhang2023enhancing,xiao2021hallucination}, while another line leverages internal states of models for detection\cite{azaria2023internal,snyder2024early,kadavath2022language}. Complementary to these approaches, several studies construct annotated hallucination datasets and train detectors on them\cite{mishra2024fine,varshney2023stitch,yang2023new}. 
On the mitigation side, existing methods aim to improve faithfulness by intervening at different stages of the generation pipeline, including model editing\cite{gao2025h,ji2025calibrating} or fine-tuning\cite{wu2024reft}, decoding corrections\cite{rebuffel2022controlling,chuangdola} and re-ranking\cite{gu2024anah}. Another practical direction reduces hallucinations via abstention or controlled stopping, encouraging models to withhold answers when uncertain\cite{tomani2024uncertainty,feng2024don}. In contrast to mitigation approaches that rely on additional training, prompt engineering, or sampling-based verification, our method focuses on activation-space interventions to directly suppress hallucination-related directions during inference, enabling risk-aware control without additional fine-tuning.

\subsection{Verifiability of AI-generated Content}

As the generative capabilities of artificial intelligence rapidly advance, models increasingly produce highly fluent and authoritative-sounding outputs. However, this fluency often masks underlying factual inaccuracies, significantly increasing the cognitive burden of human verification \cite{ji2023survey}.

In the text domain, researchers have addressed this by developing frameworks for verifiable text generation, such as training models to generate citations\cite{gao2023enabling} or utilizing self-reflection mechanisms \cite{sun2024towards}. However, in the multimodal domain, verifiability poses unique challenges. While salient factual conflicts—such as fabricating non-existent primary objects—are easily detected by users at a quick glance, fine-grained semantic deviations like incorrect textures, subtle spatial misalignments, or obscured OCR errors are notoriously difficult to verify. Works like HallusionBench \cite{guan2024hallusionbench} emphasize that these entangled visual illusions and language hallucinations easily deceive human evaluators. Despite this, prevalent multimodal hallucination benchmarks, such as POPE \cite{li2023evaluating} and MME \cite{fumme}, treat hallucinations as a binary metric, largely overlooking the varying degrees of human detectability.

Our work bridges this critical gap. By investigating the verifiability of MLLM outputs, we categorize hallucinations into obvious and elusive types. Unlike prior mitigation strategies like Woodpecker \cite{yin2024woodpecker} that apply generic corrections to all non-factual tokens, our approach leverages this detectability taxonomy to perform risk-oriented interventions in the model's activation space.
\section{Conclusion}
In this paper, we study multimodal hallucinations from a user-centric perspective and argue that hallucinations should be considered not only by whether they are incorrect, but also by how easily they can be verified by humans. a human-centered dataset from 4,470 human responses, and categorize hallucinations into obvious and elusive types according to their verifiability.

Building on this dataset, we propose an activation-space intervention method that learns separate probes for the two hallucination types and enables targeted directional ablation during inference.
Experimental results across multiple MLLMs show that the two types of hallucinations are associated with different intervention directions, and that type-matched interventions achieve stronger regulation of the corresponding verifiability. 
Moreover, we show that mixing the two intervention directions provides a flexible way to steer hallucination verifiability under different risk and usability requirements, while largely preserving general model capability under appropriate intervention strengths.

\textbf{Overall, our results suggest that hallucination mitigation in MLLMs should move beyond a binary notion of correctness and incorporate human verifiability as an optimization objective.} 
We hope this work can motivate future research on user-centered evaluation and controllable safety mechanisms for multimodal AI.

\paragraph{Limitations}
Several limitations in the current study highlight promising avenues for future research. 
On the one hand, our focus lies in the stratification and control of hallucination verifiability rather than establishing a new state-of-the-art mitigation algorithm. 
A compelling direction for future work is the integration of our verifiability-aware framework with more specially designed sophisticated techniques.
On the other hand, the current study is conducted based on an image-text verification task.
Hence, it is promising to extend to broader scenarios, such as video-based and safety-related applications.


\clearpage

\bibliographystyle{plainnat}
\bibliography{main}

\clearpage
\newpage
\beginappendix


\section{Prompts}
\label{app:prompts}

\subsection{Prompt Construction for Description}
\label{app:prompt construction}
Detailed prompt template for exhaustive description and forced inference.

\begin{tcolorbox}[fontupper=\small, fontlower=\footnotesize]
Please act as a visual detective with eagle-eyed perception. Your task is to describe every detail in this image as exhaustively as possible.

Be sure to include information from the following three dimensions. If certain parts of the image are unclear due to lighting or viewing angle, make bold inferences and fill in the missing details based on scene common sense and contextual logic. Do not answer with “unclear” or “I don’t know”:

Panoramic Objects: Describe not only the main foreground objects, but also carefully list tiny items that may appear in the background or in the corners (such as clutter on a table, decorations on a wall, etc.).

Microscopic Attributes: Precisely describe the material of objects (for example, whether something is genuine leather or synthetic leather), specific color shades, brand and model, fine-grained text or numbers appearing on objects, and the exact quantity of objects.

Spatial Interactions: Accurately describe the relative positions between objects (such as “front left of” or “partially occluded by”), the direction of people’s gaze, and any possible physical contact or causal relationships among objects.

Please begin your analysis. Make sure the description is as specific and vivid as possible, constructing a complete scene.
\end{tcolorbox}

\subsection{Prompt for Evaluation}
\label{app:prompt evaluation}
Detailed prompt template for evaluation.
\begin{tcolorbox}[fontupper=\small, fontlower=\footnotesize]
Please carefully examine the image and determine whether the following description is consistent with its content.

Description: \{Description\}

Please make your judgment strictly based on the image content, and answer only with “Yes,” “No,” or “Uncertain.”
\end{tcolorbox}

\section{Direction Selection}
\label{app:Direction Selection}

\subsection{Direction Selection Algorithm}

Given a set of difference-in-means vectors for a hallucination type $\text{type} \in \{\text{oh}, \text{eh}\}$, our goal is to select the best vector $\mathbf{r}$. Here, obvious and elusive hallucinations are processed independently, i.e., we run the same selection procedure separately on the obvious hallucination set and the elusive hallucination set. All candidate directions are selected on the validation sets only.

For each candidate vector $\mathbf{r}_i^{(l)}(\text{type})$, we compute the following::
\begin{itemize}
    \item $\mathrm{hr\_h\_score}$: the mean hallucination log-score on the hallucinated validation split $\mathcal{D}_{\text{type}}^{(\mathrm{val})}$ after ablating $\mathbf{r}_{i}^{(l)}(\text{type})$.
    \item $\mathrm{acc\_nh\_score}$: the mean accuracy log-score on the non-hallucinated validation split $\mathcal{D}_{\text{nh}}^{(\mathrm{val})}$ after the same ablation.
    \item $\mathrm{kl\_score}$: the mean KL divergence between the baseline output distribution and the ablated output distribution on $\mathcal{D}_{\text{nh}}^{(\mathrm{val})}$.
\end{itemize}

We then select the final direction by minimizing $\mathrm{hr\_h\_score}$ subject to:
\begin{itemize}
    \item $\mathrm{l < 0.9L}$: excludes the last 10\% of layers, as interventions in layers closer to the final output mapping tend to introduce larger side effects and KL divergence.
    \item $\mathrm{kl\_score < 0.1}$: filters out directions whose ablation causes excessive distribution shift on non-hallucinated examples.
    \item $\mathrm{\Delta acc\_nh < 0.1}$: ensures that the intervention does not substantially impair the model's normal capability by constraining the degradation of non-hallucinated accuracy relative to the baseline, where
\begin{equation}
    \Delta \mathrm{acc}_{\mathrm{nh}}
    = \mathrm{acc}_{\mathrm{nh}}^{\mathrm{base}}
    - \mathrm{acc}_{\mathrm{nh}}^{(i,l)}.
\end{equation}
\end{itemize}

Among all candidates that satisfy the above conditions, we select the one with the lowest hr\_h\_score on the corresponding validation set as the final direction. If no candidate passes the filtering stage, we select the direction with the minimum hr\_h\_score over the full candidate set.

\subsection{Direction Selection for Each Model}
\label{app:Direction Selection for Each Model}

We report details of direction selection for each model in Table~\ref{tab:direction_selection_details}, including the selected post-instruction token position $i$ and layer $l$ for each directions.

\begin{table*}[t]
\centering
\caption{Direction selection details for each model. Note that $i = -1$ indicates that the direction is selected from the last token position, $i = -2$ the second-to-last token position, and so on. Also note that the layer index $l$ starts from index 0, while $L$ indicates the total number of layers.}
\label{tab:direction_selection_details}

\renewcommand{\arraystretch}{1.2}
\begin{tabular}{llccccc}
\toprule
\multicolumn{1}{c}{\textbf{Model}} & \textbf{Type} & $i$ & $l / L$ & $\mathbf{\mathrm{hr\_h\_score}}$ & $\mathbf{\mathrm{\Delta acc\_nh}}$ & $\mathbf{\mathrm{kl\_score}}$ \\
\midrule

\multirow{2}{*}{Qwen2.5-VL-3B}
& \textcolor{ohicolor}{OHI} & \textcolor{ohicolor}{-1} & \textcolor{ohicolor}{29/36} & \textcolor{ohicolor}{-1.01} & \textcolor{ohicolor}{-0.32} & \textcolor{ohicolor}{0.08} \\
& \textcolor{ehicolor}{EHI} & \textcolor{ehicolor}{-2} & \textcolor{ehicolor}{25/36} & \textcolor{ehicolor}{-0.27} & \textcolor{ehicolor}{-0.08} & \textcolor{ehicolor}{0.04} \\
\midrule

\multirow{2}{*}{Qwen2.5-VL-7B}
& \textcolor{ohicolor}{OHI} & \textcolor{ohicolor}{-4} & \textcolor{ohicolor}{20/28} & \textcolor{ohicolor}{-0.54} & \textcolor{ohicolor}{-0.06} & \textcolor{ohicolor}{0.04} \\
& \textcolor{ehicolor}{EHI} & \textcolor{ehicolor}{-5} & \textcolor{ehicolor}{20/28} & \textcolor{ehicolor}{-0.49} & \textcolor{ehicolor}{0.03} & \textcolor{ehicolor}{0.04} \\
\midrule

\multirow{2}{*}{LLaVA-OneVision-1.5-8B}
& \textcolor{ohicolor}{OHI} & \textcolor{ohicolor}{-2} & \textcolor{ohicolor}{26/36} & \textcolor{ohicolor}{-1.23} & \textcolor{ohicolor}{0.02} & \textcolor{ohicolor}{0.09} \\
& \textcolor{ehicolor}{EHI} & \textcolor{ehicolor}{-2} & \textcolor{ehicolor}{23/36} & \textcolor{ehicolor}{-0.64} & \textcolor{ehicolor}{0.07} & \textcolor{ehicolor}{0.06} \\

\bottomrule
\end{tabular}
\end{table*}

\subsection{Chat Templates}
\label{app:Chat Templates}
We use the default chat template for each model.

\begin{tcolorbox}[fontupper=\small, fontlower=\footnotesize]
\textbf{Models:}\\
Qwen2.5-VL-3B\\
Qwen2.5-VL-7B\\
LLaVA-OneVision-1.5-8B\\
\textbf{Chat Template:}\\
\texttt{<|im\_start|>user\textbackslash n}\\
\texttt{<|vision\_start|><|image\_pad|><|vision\_end|>\{instruction\}\textcolor{red}{<|im\_end|>\textbackslash n}}\\
\texttt{\textcolor{red}{<|im\_start|>assistant\textbackslash n}}
\end{tcolorbox}

\section{Intervention Details and Full Results}

\subsection{Full Intervention Results}
\label{app:intervention table}

\begin{table*}[t]
\centering
\small
\setlength{\tabcolsep}{4pt}
\caption{Full intervention results across all test sets. 
For each metric, we report the absolute change in hallucination rate (HR), accuracy (ACC), and uncertain tendency (UT) under the OHI and EHI.}
\label{tab:full_intervention_results}
\resizebox{\textwidth}{!}{
\renewcommand{\arraystretch}{1.2}
\begin{tabular}{ll ccc ccc ccc}
\toprule
\multicolumn{1}{c}{\multirow{2.5}{*}{\textbf{Model}}} &
\multicolumn{1}{c}{\multirow{2.5}{*}{\textbf{Type}}} & 
\multicolumn{3}{c}{\textbf{Obvious Hallucination Subset}} & 
\multicolumn{3}{c}{\textbf{Elusive Hallucination Subset}} & 
\multicolumn{3}{c}{\textbf{Non-Hallucination Subset}} \\
\cmidrule(lr){3-5} \cmidrule(lr){6-8} \cmidrule(lr){9-11}
& & \textbf{HR(\%)} & \textbf{ACC(\%)} & \textbf{UT(\%)} 
& \textbf{HR(\%)} & \textbf{ACC(\%)} & \textbf{UT(\%)} 
& \textbf{HR(\%)} & \textbf{ACC(\%)} & \textbf{UT(\%)} \\
\midrule

\multirow{2}{*}{Qwen2.5-VL-3B}
& \textcolor{ohicolor}{OHI} & \textcolor{ohicolor}{-32.07} & \textcolor{ohicolor}{+5.79}  & \textcolor{ohicolor}{+26.28} & \textcolor{ohicolor}{-25.04} & \textcolor{ohicolor}{+4.73}  & \textcolor{ohicolor}{+20.30} & \textcolor{ohicolor}{-3.26} & \textcolor{ohicolor}{-0.64} & \textcolor{ohicolor}{+3.90} \\
& \textcolor{ehicolor}{EHI} & \textcolor{ehicolor}{-19.56} & \textcolor{ehicolor}{+2.52}  & \textcolor{ehicolor}{+17.10} & \textcolor{ehicolor}{-16.65} & \textcolor{ehicolor}{+3.15}  & \textcolor{ehicolor}{+13.50} & \textcolor{ehicolor}{-1.26} & \textcolor{ehicolor}{-3.66} & \textcolor{ehicolor}{+4.92} \\
\midrule

\multirow{2}{*}{Qwen2.5-VL-7B}
& \textcolor{ohicolor}{OHI} & \textcolor{ohicolor}{-8.06}  & \textcolor{ohicolor}{+9.14}  & \textcolor{ohicolor}{-1.08}  & \textcolor{ohicolor}{-4.76}  & \textcolor{ohicolor}{+5.22}  & \textcolor{ohicolor}{-0.47}  & \textcolor{ohicolor}{-3.32} & \textcolor{ohicolor}{+5.74} & \textcolor{ohicolor}{-2.43} \\
& \textcolor{ehicolor}{EHI} & \textcolor{ehicolor}{-17.42} & \textcolor{ehicolor}{+16.99} & \textcolor{ehicolor}{+0.43}  & \textcolor{ehicolor}{-15.08} & \textcolor{ehicolor}{+14.22} & \textcolor{ehicolor}{+0.85}  & \textcolor{ehicolor}{+2.72} & \textcolor{ehicolor}{-1.78} & \textcolor{ehicolor}{-0.94} \\
\midrule

\multirow{2}{*}{LLaVA-OneVision-1.5-8B}
& \textcolor{ohicolor}{OHI} & \textcolor{ohicolor}{-35.26} & \textcolor{ohicolor}{+32.75} & \textcolor{ohicolor}{+2.51}  & \textcolor{ohicolor}{-26.58} & \textcolor{ohicolor}{+24.82} & \textcolor{ohicolor}{+1.76}  & \textcolor{ohicolor}{-1.78} & \textcolor{ohicolor}{+1.39} & \textcolor{ohicolor}{+0.39} \\
& \textcolor{ehicolor}{EHI} & \textcolor{ehicolor}{-17.53} & \textcolor{ehicolor}{+18.51} & \textcolor{ehicolor}{-0.88}  & \textcolor{ehicolor}{-19.43} & \textcolor{ehicolor}{+19.75} & \textcolor{ehicolor}{+0.32}  & \textcolor{ehicolor}{+3.99} & \textcolor{ehicolor}{-3.27} & \textcolor{ehicolor}{-0.71} \\
\bottomrule
\end{tabular}
}
\end{table*}

For completeness, we provide the full intervention results across all three test sets in Table~\ref{tab:full_intervention_results}. 

Consistent with the findings in the main paper, probe-based intervention generally reduces hallucination rate and improves accuracy on both the Obvious and Elusive Hallucination test sets. 
On the Non-Hallucination test set, the performance variations are comparatively small, suggesting that the intervention mainly affects hallucination-prone cases while introducing only limited side effects on faithful samples. 

In addition, the selected intervention coefficients for each model are reported in Appendix~\ref{app:intervention coefficients}.

\subsection{Sensitivity to Intervention Coefficient on General Ability}
\label{app:alpha_sweep}
\begin{figure*}[!t]
    \centering
    \begin{minipage}[t]{0.48\textwidth}
        \centering
        \includegraphics[width=\textwidth]{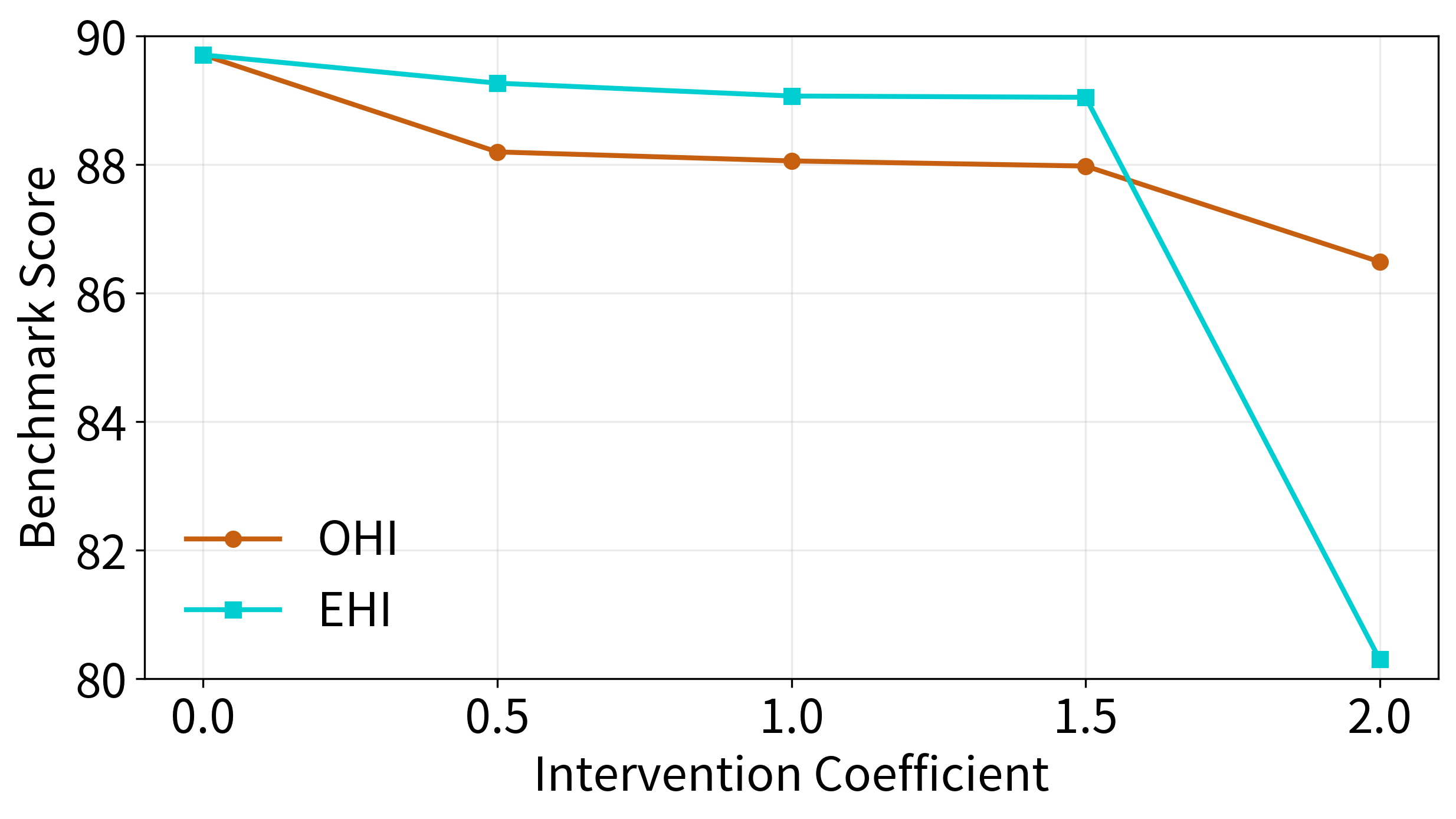}
        \caption{TextVQA accuracy of Qwen2.5-VL-7B under OHI and EHI as a function of the intervention coefficient.}
        \label{figs:alpha_sweep}
    \end{minipage}
    \hfill
    \begin{minipage}[t]{0.48\textwidth}
        \centering
        \includegraphics[width=\textwidth]{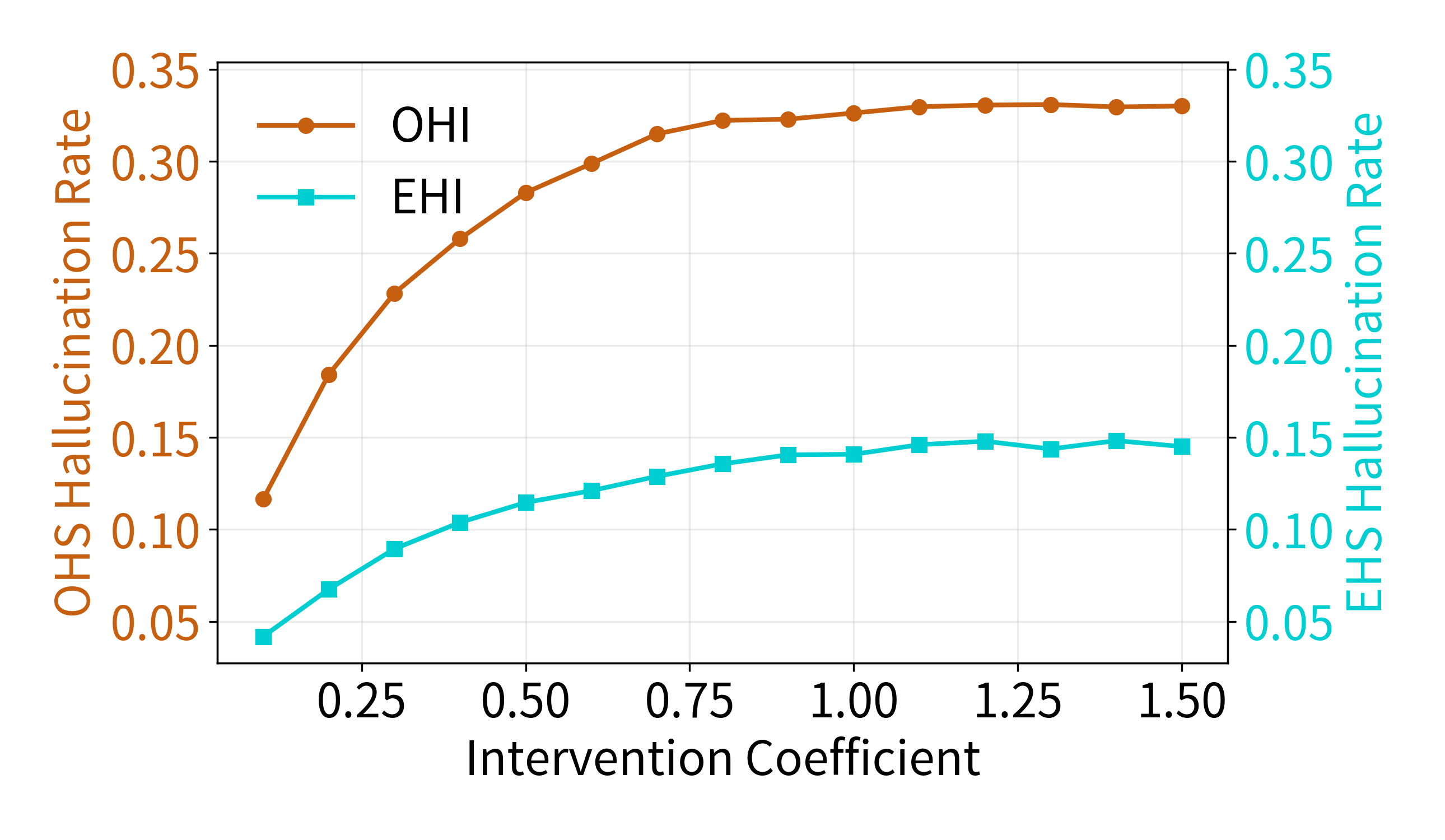}
        \caption{Validation-based sweep of intervention coefficients on Qwen2.5-VL-3B. 
        The left y-axis corresponds to the performance of OHI evaluated on the OHS, 
        while the right y-axis corresponds to the performance of EHI evaluated on the EHS.}
        \label{figs:intervention_coefficients}
    \end{minipage}
\end{figure*}

To further characterize sensitivity to intervention coefficient, we sweep the intervention coefficient $\alpha$ on TextVQA using Qwen2.5-VL-7B, as shown in Figure~\ref{figs:alpha_sweep}.
Accuracy remains largely stable within a moderate range of $\alpha$, while large $\alpha$ values can lead to sharp performance degradation. This suggests that over-ablation may adversely affect general model ability, further motivating our use of moderate, validation-selected intervention coefficients.

\subsection{Intervention Coefficients}
\label{app:intervention coefficients}

We select the intervention coefficients on the validation sets by sweeping the coefficient $\alpha$ separately for OHI and EHI. 
We vary the OHI coefficient $\alpha_{\text{oh}}$ on the obvious validation set and the EHI coefficient $\alpha_{\text{eh}}$ on the elusive validation set, as illustrated in Figure~\ref{figs:intervention_coefficients}.


To reduce potential side effects on the model, we restrict the selected intervention strength to approximately $\alpha < 1.5$. This choice is supported by Appendix~\ref{app:alpha_sweep}, which shows that model performance remains relatively stable only within a moderate range of $\alpha$, while large coefficients can cause sharp degradation. Rather than searching for a single globally optimal coefficient, we choose a coefficient once the hallucination rate on the corresponding validation set no longer decreases substantially with larger $\alpha$. If no such point satisfies this criterion, we set $\alpha = 1$. In other words, we treat the acceptable intervention strength as a stable range rather than a unique best point.
\begin{wraptable}{r}{0.5\textwidth}
\centering
\caption{Selected intervention coefficients for OHI and EHI.}
\label{tab:intervention coefficients}
\begin{tabular}{lcc}
\toprule
\multicolumn{1}{c}{Model} & $\alpha_{\text{oh}}$ & $\alpha_{\text{eh}}$ \\
\midrule
Qwen2.5-VL-3B          & 0.90 & 0.90 \\
Qwen2.5-VL-7B          & 0.80 & 0.70 \\
LLaVA-OneVision-1.5-8B & 0.80 & 0.80 \\
\bottomrule
\end{tabular}
\end{wraptable}

For Qwen2.5-VL-3B, the validation curves in Figure~\ref{figs:intervention_coefficients} become relatively flat in the high-performing region, and we therefore select $\alpha_{\text{oh}}=0.9$ and $\alpha_{\text{eh}}=0.9$. The final selected intervention strengths for all models are summarized in Table~\ref{tab:intervention coefficients}.

\end{document}